\documentclass[11pt,a4paper]{article}
\usepackage[utf8]{inputenc}
\usepackage[T1]{fontenc}
\usepackage{graphicx}
\usepackage{booktabs}
\usepackage{amsmath}
\usepackage{amssymb}
\usepackage{hyperref}
\usepackage{geometry}
\usepackage{authblk}
\usepackage{multirow}
\usepackage{microtype}
\usepackage{xcolor}   

\title{Student Capacity Moderates Knowledge Distillation Effectiveness:\\
A Systematic Study Across ResNet Teacher-Student Pairs on CIFAR-10}

\author{Umut Onur Ya\c{s}ar}
\affil{\href{mailto:umutonuryasar@gmail.com}{umutonuryasar@gmail.com}}
\date{}

\begin{document}

\maketitle

\begin{abstract}
We investigate how teacher-student capacity relationships modulate knowledge
distillation (KD) effectiveness in ResNet-based image classification on
CIFAR-10. Across four teacher-student pairs --- R50$\!\rightarrow$R18,
R34$\!\rightarrow$R18, R50$\!\rightarrow$R34, and R101$\!\rightarrow$R34 ---
we compare Logit-KD and Feature-KD under a strict evaluation protocol:
hyperparameters and checkpoints are selected on a held-out validation split,
selected configurations are re-run with five seeds, and the test set is used
exclusively for final reporting. Beyond accuracy, we measure distillation
\emph{fidelity} directly via teacher-student agreement and KL divergence. We
report four findings. First, the student-capacity pattern survives the
corrected protocol at reduced magnitude: the only statistically significant
gains occur for R34 students under Feature-KD ($+0.19$ and $+0.21$\,pp,
$p<0.05$), in two pairs whose teachers differ two-fold in parameters but not
in accuracy --- localizing the moderating variable on the student side ---
while no KD gain for R18 students is distinguishable from zero. Second,
Feature-KD matches or outperforms Logit-KD in all four pairs, and its
students land closer to the teacher's output distribution ($D_{KL}$ at
$T{=}1$) than Logit-KD students despite never observing teacher logits.
Third, top-1 teacher-student agreement is flat across all pairs, decoupling
fidelity from accuracy gains. Fourth, architecture dominates KD: correcting
the ResNet stem for $32{\times}32$ inputs is worth $+5.5$ to $+7.2$\,pp ---
more than $25\times$ the largest KD gain. We also retract an attribution made
in v1 of this paper: a controlled re-run shows the reported gradient-clipping
bug had no measurable effect, and v1's larger gains are explained by
test-set selection. All code and results are available at
\href{https://github.com/umutonuryasar/kd-capacity-gap}{github.com/umutonuryasar/kd-capacity-gap}.
\end{abstract}

\section{Introduction}

Deep neural networks achieve state-of-the-art performance across a wide range
of computer vision tasks, but their computational demands make deployment on
resource-constrained hardware challenging. Knowledge Distillation (KD) offers a
principled compression strategy: instead of training a small student model on
hard one-hot labels alone, the student also learns from the soft output
distribution of a larger, pre-trained teacher model. These soft targets encode
inter-class similarity --- the relative probabilities assigned to non-target
classes --- a form of structured information Hinton et al.~\cite{hinton2015}
termed ``dark knowledge.''

Despite its widespread use, the relative effectiveness of different distillation
paradigms remains context-dependent and poorly understood. Two dominant
approaches exist: Logit-KD, which aligns the student's output distribution with
the teacher's via KL divergence, and Feature-KD, which aligns intermediate
representations at one or more layers. Both approaches introduce
hyperparameters --- the distillation weight $\alpha$, the softmax temperature
$T$ for Logit-KD, and layer selection and alignment metrics for Feature-KD ---
whose optimal values depend on the task and architecture.

Prior work has shown that excessive teacher-student capacity mismatch can
impede knowledge transfer \cite{cho2019efficacy,mirzadeh2020improved}, but
systematic evidence across multiple teacher-student pairs and KD paradigms
under a leakage-free evaluation protocol remains limited. Moreover, recent
work argues that distillation gains should be interpreted jointly with
\emph{fidelity} --- how closely the student actually matches its teacher ---
rather than accuracy alone \cite{stanton2021does}.

In this work, we make the following contributions:
\begin{enumerate}
    \item A systematic comparison of Logit-KD and Feature-KD on CIFAR-10
    across four teacher-student pairs under a strict two-stage protocol:
    hyperparameters ($\alpha \in \{0.3, 0.5, 0.7\}$, $T \in \{2, 3, 4\}$)
    selected on a held-out validation split, selected configurations re-run
    with five seeds, and the test set reserved for final reporting.
    \item Fidelity measurements --- teacher-student top-1 agreement,
    $D_{KL}(p_T \!\parallel\! p_S)$, and per-class accuracy --- that
    qualify the capacity interpretation: accuracy gains concentrate on
    R34 students, yet top-1 agreement is flat across pairs, while
    Feature-KD students are uniformly closer to the teacher in $D_{KL}$
    at $T{=}1$.
    \item A quantified demonstration that standard torchvision ResNet stems
    are poorly suited to $32{\times}32$ inputs
    (Table~\ref{tab:stem}), and that architectural correctness dominates KD
    gains by an order of magnitude.
    \item A transparent audit trail of two methodological corrections
    relative to the first version of this study: a gradient-clipping bug
    affecting Feature-KD projection layers, and the replacement of
    test-set-based selection with validation-based selection. We report how
    each correction changes the conclusions.
\end{enumerate}

\section{Related Work}

\subsection{Knowledge Distillation}
Hinton et al.~\cite{hinton2015} introduced KD as soft-target training,
minimizing KL divergence between temperature-scaled teacher and student
distributions. The combined loss $L = \alpha \cdot L_{kd} + (1-\alpha) \cdot
L_{ce}$ balances distillation against ground-truth supervision. Temperature $T$
controls soft target sharpness: higher $T$ exposes inter-class similarity.
Zhao et al.~\cite{zhao2022decoupled} recently decomposed the KD loss into
target-class and non-target-class components (DKD), showing that the two carry
distinct information and should be weighted separately. Huang et
al.~\cite{huang2022stronger} relaxed exact distribution matching to
correlation-based matching (DIST), showing that strict KL alignment can be
counterproductive when the teacher-student gap is large --- a finding directly
relevant to capacity analysis. Beyer et al.~\cite{beyer2022good} demonstrated
that distillation effectiveness depends strongly on training protocol
(duration, augmentation consistency), underscoring that protocol choices can
dominate method differences.

\subsection{Feature-Based Distillation}
Romero et al.~\cite{romero2015fitnets} proposed FitNets, aligning intermediate
feature maps via $L_2$ loss after linear projection.
Zagoruyko and Komodakis~\cite{zagoruyko2017paying} extended this to attention
maps, showing that spatial attention patterns transfer more effectively than raw
activations. Chen et al.~\cite{chen2021distilling} introduced ReviewKD,
aggregating features across multiple teacher layers via cross-attention,
achieving strong results on CIFAR-100 and ImageNet.
Tian et al.~\cite{tian2020contrastive} proposed CRD, framing feature
distillation as a contrastive learning objective and demonstrating consistent
improvements over MSE-based alignment. Our Feature-KD uses MSE with
$1{\times}1$ projection --- a simpler baseline intentionally chosen to isolate
the effect of capacity gap from implementation choices.

\subsection{Teacher-Student Capacity Gap and Distillation Fidelity}
A critical but underexplored question is whether a stronger teacher always
yields a better student. Cho and Hariharan~\cite{cho2019efficacy} first
systematically studied this, showing that an overly powerful teacher can harm
distillation because the student struggles to match distributions too far from
its own capacity. Mirzadeh et al.~\cite{mirzadeh2020improved} proposed Teacher
Assistant KD (TAKD), inserting an intermediate-capacity model between teacher
and student to bridge the gap. Stanton et al.~\cite{stanton2021does}
demonstrated that students often fail to match their teachers even when
distillation improves generalization, arguing that agreement and predictive
divergence should be reported alongside accuracy. Our work provides
complementary evidence: we hold the dataset and training protocol fixed, vary
teacher and student capacity across four ResNet pairs, and report fidelity
metrics for every final configuration.

\subsection{Architecture and Input Resolution}
He et al.~\cite{he2016deep} designed ResNets for ImageNet ($224{\times}224$).
The initial $7{\times}7$ convolution with stride 2 followed by MaxPool reduces
spatial resolution to $56{\times}56$ before the first residual block ---
appropriate for high-resolution inputs but destructive for $32{\times}32$ CIFAR
images, where the same operations produce an $8{\times}8$ feature map. Standard
CIFAR benchmarks replace this stem with a $3{\times}3$ convolution at stride 1
and remove MaxPool. We adopt this modification and demonstrate that
architectural correctness is a prerequisite for effective distillation.

\section{Method}

\subsection{Loss Formulation}
The total training loss follows Hinton et al.~\cite{hinton2015}:
\begin{equation}
L_{\text{total}} = \alpha \cdot L_{kd} + (1 - \alpha) \cdot L_{ce}
\end{equation}
where $L_{ce} = \text{CrossEntropy}(z_S, y)$ and $\alpha \in [0, 1]$ controls
the distillation weight.

\subsection{Logit-KD}
\begin{equation}
L_{\text{logit}} = T^2 \cdot D_{KL}\!\left(
  \text{softmax}(z_T/T) \;\|\; \text{softmax}(z_S/T)
\right)
\end{equation}
The $T^2$ factor compensates for reduced gradient magnitude at high temperature.
We note the standard observation that minimizing
$D_{KL}(p_T \!\parallel\! p_S)$ in the student parameters is equivalent to
maximum-likelihood estimation under the teacher's output distribution.

\subsection{Feature-KD}
For each layer $i \in \{\text{layer1}, \text{layer2}, \text{layer3},
\text{layer4}\}$, we compute MSE between projected student features and teacher
features, plus cosine similarity on globally averaged pooled (GAP) projected
features:
\begin{equation}
L_{\text{feat}} = \operatorname{mean}_i\!\left(
  \text{MSE}(\text{proj}_i(S_i),\, T_i)
\right)
+ \beta \cdot \operatorname{mean}_i\!\left(
  1 - \text{cos\_sim}(\text{GAP}(\text{proj}_i(S_i)),\, \text{GAP}(T_i))
\right)
\end{equation}
with $\beta=0.5$. Channel mismatches are resolved via Xavier-initialized
$1{\times}1$ Conv2d projections whose parameters are trained jointly with the
student and included in gradient clipping (see Section~\ref{sec:impl}).

\subsection{CIFAR-Specific Architecture}
We replace the standard conv1 (kernel=7, stride=2) and MaxPool with conv1
(kernel=3, stride=1) and Identity for all models. This preserves the full
$32{\times}32$ spatial resolution through the first residual block and is
applied consistently to all teacher and student variants.

\subsection{Teacher-Student Pairs}
We study four capacity configurations (parameter counts for the CIFAR-adapted
models):
\begin{itemize}
    \item \textbf{R50$\!\rightarrow$R18:} 23.5M-parameter teacher,
    11.2M-parameter student (cross-family: Bottleneck teacher, BasicBlock
    student).
    \item \textbf{R34$\!\rightarrow$R18:} 21.3M-parameter teacher,
    11.2M-parameter student (same block family, smaller capacity gap).
    \item \textbf{R50$\!\rightarrow$R34:} 23.5M-parameter teacher,
    21.3M-parameter student (larger student capacity, same teacher).
    \item \textbf{R101$\!\rightarrow$R34:} 42.5M-parameter teacher,
    21.3M-parameter student (larger student, larger teacher gap; new in v2 ---
    separates ``R34 as student'' from ``R50 as teacher'' as explanatory
    variables).
\end{itemize}

\section{Experiments}

\subsection{Evaluation Protocol}
\label{sec:protocol}
The 50k CIFAR-10 training set is partitioned once, with a fixed split seed
independent of all run seeds, into 45k training and 5k validation examples;
every run in the study shares this split. All model selection ---
hyperparameter choice \emph{and} best-epoch checkpointing --- uses validation
accuracy only. The 10k test set is evaluated exactly twice per run: once with
the best-validation checkpoint and once with the final-epoch weights; we report
the former and release both.

Experiments follow a two-stage protocol. \textbf{Stage 1 (selection):} the
full grid (Logit-KD: $\alpha \in \{0.3, 0.5, 0.7\} \times T \in \{2, 3, 4\}$;
Feature-KD: $\alpha \in \{0.3, 0.5, 0.7\}$, $\beta{=}0.5$) is run with a
single seed per pair, and the best configuration per (pair, method) is
selected by validation accuracy. \textbf{Stage 2 (final):} selected
configurations and cross-entropy baselines are re-run with five seeds
$\{0,\dots,4\}$; all reported numbers are mean\,$\pm$\,std of test accuracy
over these seeds. The complete Stage-1 grid is provided in the repository.

\subsection{Setup}
\textbf{Dataset:} CIFAR-10 (45k train / 5k val / 10k test after the split of
Section~\ref{sec:protocol}, 10 classes, $32{\times}32$).
\textbf{Augmentation:} RandomCrop(32, padding=4), RandomHorizontalFlip,
Normalize($\mu$=[0.4914, 0.4822, 0.4465], $\sigma$=[0.2470, 0.2435, 0.2616]).
\textbf{Optimizer:} SGD (momentum=0.9, weight\_decay=$5{\times}10^{-4}$,
Nesterov=True). \textbf{LR:} 0.1 with CosineAnnealingLR ($T_{\max}=100$,
$\eta_{\min}=10^{-4}$) for students; teachers trained for 200 epochs.
\textbf{Batch size:} 128. \textbf{Hardware:} NVIDIA A100.
\textbf{Reproducibility:} deterministic cuDNN; seeded Python, NumPy, and
Torch RNGs; seeded DataLoader workers. Teachers are trained with three seeds
and their variance reported; the seed-0 best-validation checkpoint serves as
the canonical teacher for all KD runs.

\subsection{Implementation Details}
\label{sec:impl}
\textbf{Gradient clipping:} We apply \texttt{clip\_grad\_norm\_} with
threshold 1.0 to the union of student model parameters \emph{and} Feature-KD
projection layer parameters. This distinction matters: excluding projection
layers leads to unclipped gradient norms of up to 4.65 in early training,
causing optimization instability. We identified this bug by auditing an
earlier implementation (corresponding to our CS229 course paper, single-seed,
R50$\!\rightarrow$R18 only).
Table~\ref{tab:bug} reports a controlled re-run of the bugged variant under
the v2 protocol: identical split, baseline, $\alpha$, and seeds as the
corrected Stage-2 runs, differing only in whether projection parameters are
included in gradient clipping. The two variants are statistically
indistinguishable ($p{=}0.69$), and the recorded pre-clip projection gradient
norms never exceed $0.21$ --- a fifth of the clipping threshold. The clipping
operation is therefore inert for these layers in this setting, and we retract
v1's attribution of Feature-KD underperformance to this bug: the apparent v1
effect is explained by single-seed variance and test-set selection
(Section~\ref{sec:results}). We could not reproduce v1's reported unclipped
norm of 4.65 under the v2 protocol.

\begin{table}[h]
\centering
\caption{Bugged (projections excluded from clipping) vs.\ corrected
Feature-KD for R50$\!\rightarrow$R18 under identical v2 conditions
($\alpha{=}0.7$, 5 seeds).}
\label{tab:bug}
\vspace{2mm}
\begin{tabular}{lccc}
\toprule
Variant & Test Acc & Max proj.\ grad norm & Welch $p$ \\
\midrule
Corrected (proj.\ clipped) & 94.95\% $\pm$0.20 & --- (clipped at 1.0) &
\multirow{2}{*}{0.69} \\
Bugged (proj.\ unclipped)  & 95.00\% $\pm$0.18 & 0.21 & \\
\bottomrule
\end{tabular}
\end{table}

\textbf{Teacher loading:} Missing teacher weights raise a hard
\texttt{ValueError}; silent fallback to a random teacher is not permitted.

\textbf{Fidelity metrics:} For every Stage-2 run we compute teacher-student
top-1 agreement and $D_{KL}(p_T \!\parallel\! p_S)$ on the test set at
$T \in \{1, 4\}$, plus per-class student accuracy \cite{stanton2021does}.

\paragraph{Changes from v1.}
The first version of this paper selected hyperparameters and checkpoints on
the test set, used three seeds, single-seed teachers, and three pairs.
Version 2 introduces the validation-based protocol of
Section~\ref{sec:protocol}, five-seed final runs, three-seed teachers, a
fourth pair (R101$\!\rightarrow$R34), fidelity metrics, and a measured stem
ablation. Where v1 and v2 conclusions differ, we attribute the difference
explicitly.

\section{Results}
\label{sec:results}

\subsection{Teacher and Baseline Accuracy}
Table~\ref{tab:teachers} reports teacher and baseline accuracies under the v2
protocol.

\begin{table}[h]
\centering
\caption{Teacher (3 seeds) and baseline (5 seeds) test accuracies
(CIFAR-specific architecture; best-validation checkpoints).}
\label{tab:teachers}
\vspace{2mm}
\begin{tabular}{lcc}
\toprule
Model & Accuracy & Std \\
\midrule
Teacher ResNet-101 & 95.37\% & $\pm$0.12 \\
Teacher ResNet-50  & 95.36\% & $\pm$0.18 \\
Teacher ResNet-34  & 95.30\% & $\pm$0.09 \\
Baseline ResNet-18 & 94.86\% & $\pm$0.14 \\
Baseline ResNet-34 & 95.04\% & $\pm$0.13 \\
\bottomrule
\end{tabular}
\end{table}

\subsection{Distillation Results Across Capacity Pairs}

Tables~\ref{tab:r50r18}--\ref{tab:r101r34} report distillation results per
pair. $\Delta$\,Baseline denotes mean accuracy minus the corresponding student
baseline; Agree denotes teacher-student top-1 agreement on the test set; KL
denotes $D_{KL}(p_T \!\parallel\! p_S)$ at $T{=}4$.

\begin{table}[h]
\centering
\caption{R50$\!\rightarrow$R18 distillation results (baseline: 94.86\%).
\checkmark\,=\,best configuration per pair. Best hyperparameters (selected on
val): Logit-KD $\alpha{=}0.5$, $T{=}4$; Feature-KD $\alpha{=}0.7$.}
\label{tab:r50r18}
\vspace{2mm}
\begin{tabular}{lccccc}
\toprule
Config & Mean Acc & Std & $\Delta$ Baseline & Agree & KL@4 \\
\midrule
Teacher (R50)  & 95.36\% & $\pm$0.18 & --- & --- & --- \\
Baseline (R18) & 94.86\% & $\pm$0.14 & --- & --- & --- \\
\midrule
Logit-KD   & 94.91\% & $\pm$0.08 & $+0.06$\,pp & 95.33\% & 0.028 \\
Feature-KD\,\checkmark & 94.95\% & $\pm$0.20 & $+0.10$\,pp & 95.49\% & 0.037 \\
\bottomrule
\end{tabular}
\end{table}

\begin{table}[h]
\centering
\caption{R34$\!\rightarrow$R18 distillation results (baseline: 94.86\%).
Best hyperparameters: Logit-KD $\alpha{=}0.7$, $T{=}2$; Feature-KD
$\alpha{=}0.7$.}
\label{tab:r34r18}
\vspace{2mm}
\begin{tabular}{lccccc}
\toprule
Config & Mean Acc & Std & $\Delta$ Baseline & Agree & KL@4 \\
\midrule
Teacher (R34)  & 95.30\% & $\pm$0.09 & --- & --- & --- \\
Baseline (R18) & 94.86\% & $\pm$0.14 & --- & --- & --- \\
\midrule
Logit-KD   & 94.78\% & $\pm$0.09 & $-0.08$\,pp & 95.50\% & 0.024 \\
Feature-KD\,\checkmark & 94.98\% & $\pm$0.15 & $+0.13$\,pp & 95.53\% & 0.029 \\
\bottomrule
\end{tabular}
\end{table}

\begin{table}[h]
\centering
\caption{R50$\!\rightarrow$R34 distillation results (baseline: 95.04\%).
Best hyperparameters: Logit-KD $\alpha{=}0.3$, $T{=}3$; Feature-KD
$\alpha{=}0.5$. $^{*}$\,$p<0.05$ (Welch) vs.\ baseline.}
\label{tab:r50r34}
\vspace{2mm}
\begin{tabular}{lccccc}
\toprule
Config & Mean Acc & Std & $\Delta$ Baseline & Agree & KL@4 \\
\midrule
Teacher (R50)  & 95.36\% & $\pm$0.18 & --- & --- & --- \\
Baseline (R34) & 95.04\% & $\pm$0.13 & --- & --- & --- \\
\midrule
Logit-KD   & 95.13\% & $\pm$0.16 & $+0.08$\,pp & 95.39\% & 0.031 \\
Feature-KD\,\checkmark & 95.24\% & $\pm$0.10 & $+0.19$\,pp$^{*}$ & 95.41\% & 0.028 \\
\bottomrule
\end{tabular}
\end{table}

\begin{table}[h]
\centering
\caption{R101$\!\rightarrow$R34 distillation results (baseline: 95.04\%).
Best hyperparameters: Logit-KD $\alpha{=}0.7$, $T{=}2$; Feature-KD
$\alpha{=}0.3$. $^{*}$\,$p<0.05$ (Welch) vs.\ baseline.}
\label{tab:r101r34}
\vspace{2mm}
\begin{tabular}{lccccc}
\toprule
Config & Mean Acc & Std & $\Delta$ Baseline & Agree & KL@4 \\
\midrule
Teacher (R101) & 95.37\% & $\pm$0.12 & --- & --- & --- \\
Baseline (R34) & 95.04\% & $\pm$0.13 & --- & --- & --- \\
\midrule
Logit-KD   & 95.10\% & $\pm$0.16 & $+0.06$\,pp & 95.50\% & 0.032 \\
Feature-KD\,\checkmark & 95.26\% & $\pm$0.11 & $+0.21$\,pp$^{*}$ & 95.53\% & 0.030 \\
\bottomrule
\end{tabular}
\end{table}

\subsection{Capacity Gap Summary}
Table~\ref{tab:summary} summarises KD gains and teacher-student agreement
across all pairs. The accuracy pattern of v1 replicates directionally at
reduced magnitude: both statistically significant gains belong to R34
students under Feature-KD, while top-1 agreement is statistically flat
across all cells.

\begin{table}[h]
\centering
\caption{Summary of KD gains ($\Delta$ Baseline, pp) and teacher-student
top-1 agreement across all pairs. $^{*}$\,$p<0.05$ (Welch) vs.\ baseline;
no single comparison survives Bonferroni correction across the eight tests
($\alpha{=}0.00625$).}
\label{tab:summary}
\vspace{2mm}
\begin{tabular}{llccccc}
\toprule
Teacher & Student & T-S Gap & Logit $\Delta$ & Feature $\Delta$ & Best &
Agree (best) \\
\midrule
R34 (95.30\%)  & R18 (94.86\%) & 0.44\,pp & $-0.08$ & $+0.13$ & Feature & 95.53\% \\
R50 (95.36\%)  & R18 (94.86\%) & 0.50\,pp & $+0.06$ & $+0.10$ & Feature & 95.49\% \\
R50 (95.36\%)  & R34 (95.04\%) & 0.32\,pp & $+0.08$ & $+0.19^{*}$ & Feature & 95.41\% \\
R101 (95.37\%) & R34 (95.04\%) & 0.33\,pp & $+0.06$ & $+0.21^{*}$ & Feature & 95.53\% \\
\bottomrule
\end{tabular}
\end{table}

\subsection{Stem Ablation}
Table~\ref{tab:stem} quantifies the effect of the input stem. Replacing the
ImageNet stem ($7{\times}7$ convolution, stride 2, MaxPool) with the
CIFAR-specific stem ($3{\times}3$, stride 1, no MaxPool) is worth $+5.50$ to
$+7.15$\,pp, with the largest effect on the smallest model (R18). Every KD
gain in this study is at least an order of magnitude smaller than this
architectural correction.

\begin{table}[h]
\centering
\caption{ImageNet stem ($7{\times}7$, stride 2, MaxPool) vs.\ CIFAR stem
($3{\times}3$, stride 1, no MaxPool). Students: 3 seeds; R50: single seed.}
\label{tab:stem}
\vspace{2mm}
\begin{tabular}{lccc}
\toprule
Model & ImageNet stem & CIFAR stem & $\Delta$ (pp) \\
\midrule
ResNet-18 & 87.71\% $\pm$0.22 & 94.86\% $\pm$0.14 & $+7.15$ \\
ResNet-34 & 88.28\% $\pm$0.06 & 95.02\% $\pm$0.14 & $+6.74$ \\
ResNet-50 & 89.86\% & 95.36\% & $+5.50$ \\
\bottomrule
\end{tabular}
\end{table}

\begin{figure}[t]
\centering
\includegraphics[width=0.75\textwidth]{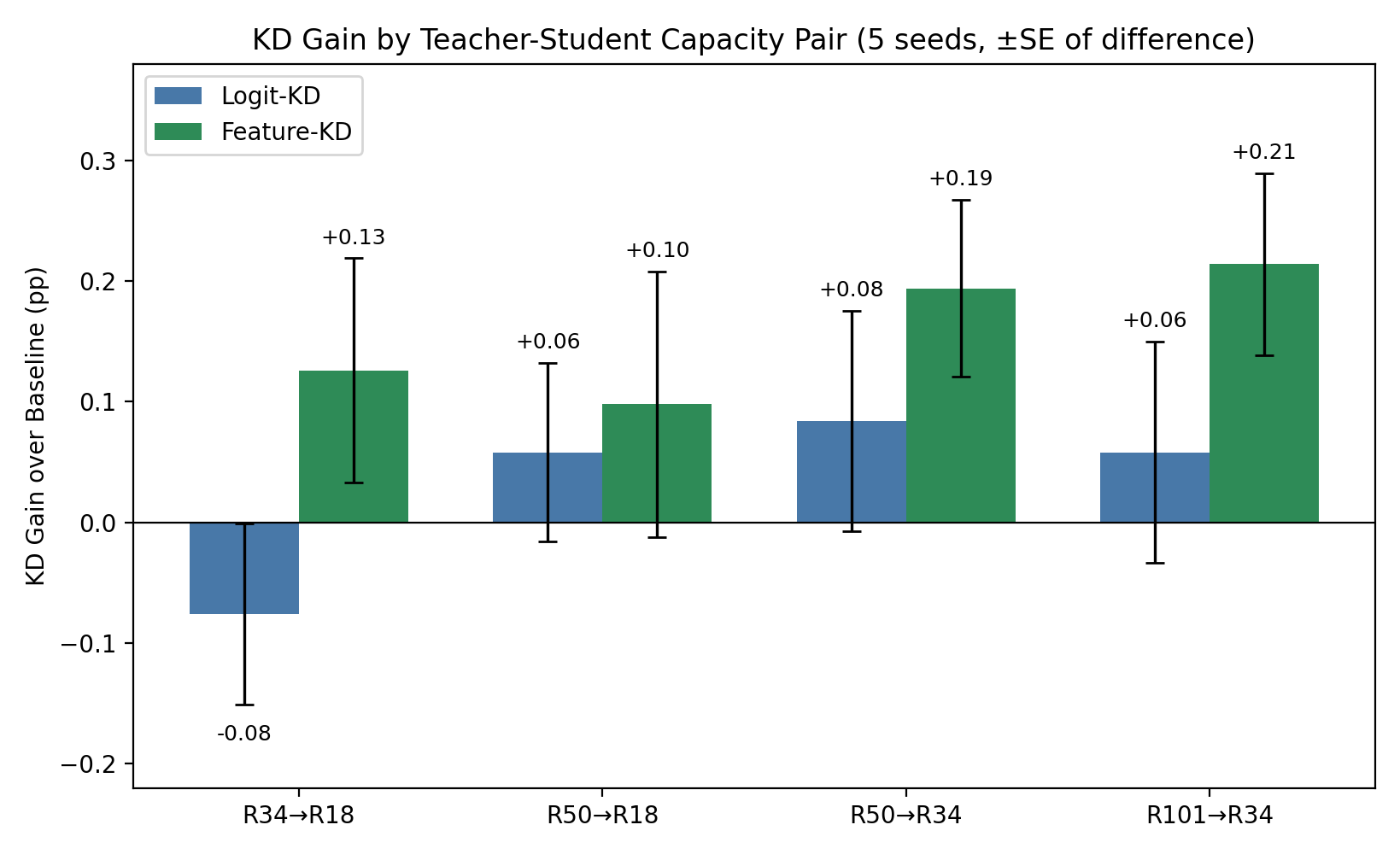}
\caption{KD gain over baseline (pp) across four teacher-student pairs
(5 seeds; error bars show the standard error of the difference vs.\
baseline). Both statistically significant gains (R50$\!\rightarrow$R34
and R101$\!\rightarrow$R34 Feature-KD) belong to R34 students.}
\label{fig:capacity_gap}
\end{figure}

\begin{figure}[t]
\centering
\includegraphics[width=\textwidth]{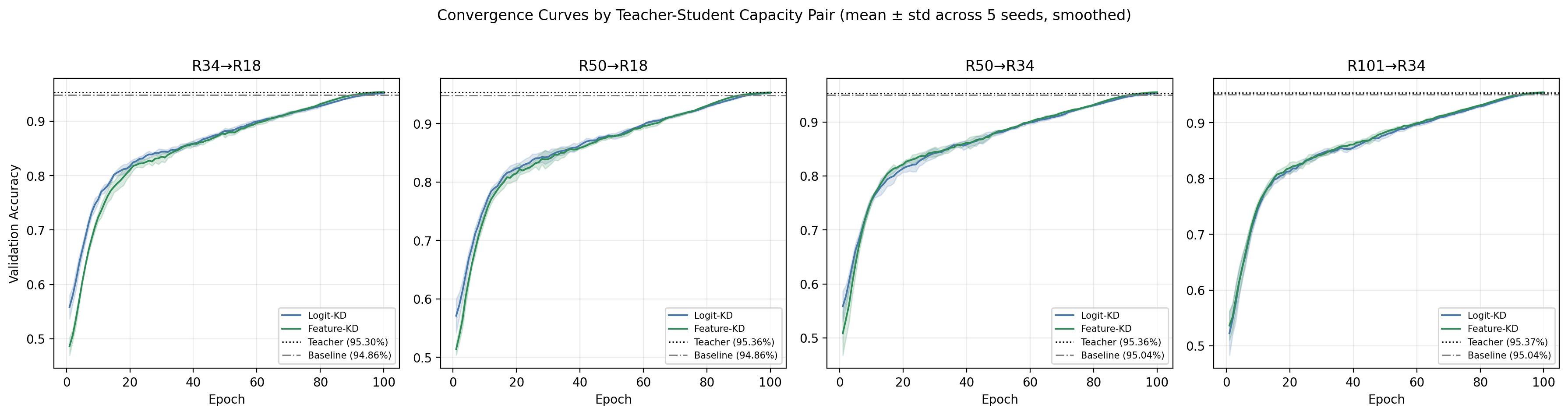}
\caption{Validation accuracy curves for all four pairs (mean $\pm$ std
across 5 seeds, EMA smoothed). Dotted line = teacher accuracy; dash-dot
line = baseline accuracy.}
\label{fig:convergence}
\end{figure}

\section{Discussion}

\paragraph{Student Capacity as the Locus of Moderation.}
Across all four pairs, the only statistically significant KD gains belong to
R34 students under Feature-KD: R50$\!\rightarrow$R34 ($+0.19$\,pp,
$p{=}0.031$) and R101$\!\rightarrow$R34 ($+0.21$\,pp, $p{=}0.022$), while no
gain for an R18 student is distinguishable from zero
(Table~\ref{tab:summary}). The R101$\!\rightarrow$R34 pair, added in this
version, is what localizes the effect: R101 and R50 differ almost two-fold in
parameters (42.5M vs.\ 23.5M) yet are statistically indistinguishable in
accuracy (95.37\% vs.\ 95.36\%), and the R34 student's gain is essentially
unchanged between them. Neither teacher accuracy gap --- which is, if
anything, \emph{smaller} for the R34 pairs (0.32--0.33\,pp vs.\
0.44--0.50\,pp) --- nor teacher size predicts the gain; student identity
does. This pattern is consistent with student representational capacity
moderating how much a student can extract from distillation, in line with
capacity-mismatch accounts \cite{cho2019efficacy,mirzadeh2020improved}. Two
caveats bound this claim: student capacity remains partially confounded with
block family (both R34 pairs use a BasicBlock student, but so does every R18
pair), and only two student sizes are examined.

\paragraph{Logit-KD vs.\ Feature-KD.}
Feature-KD matches or outperforms Logit-KD in all four pairs
(Figure~\ref{fig:capacity_gap}), with the largest margin on the same-family
pair R34$\!\rightarrow$R18 ($+0.20$\,pp difference) and the smallest on the
cross-family pair R50$\!\rightarrow$R18 ($+0.04$\,pp). This reverses the
conclusion of the first version of this study, which reported Logit-KD
consistently ahead; as Section~\ref{sec:impl} shows, that conclusion was an
artifact of single-seed evaluation and test-set selection rather than of the
gradient-clipping bug it was attributed to. We emphasise the scope of the
present comparison: our Feature-KD is a deliberately simple MSE-plus-cosine
baseline, and these results say nothing about stronger logit-based methods
such as DKD \cite{zhao2022decoupled} or DIST \cite{huang2022stronger}.

\paragraph{The R34$\!\rightarrow$R18 Logit-KD Anomaly Replicates.}
In v1, R34$\!\rightarrow$R18 Logit-KD was the sole configuration with no
gain ($+0.00$\,pp); under the corrected protocol it is again the weakest cell
($-0.08$\,pp, $p{=}0.35$, i.e.\ consistent with zero), while Feature-KD on
the same pair is positive ($+0.13$\,pp). The explanation offered in v1
remains plausible: with a same-family teacher only 0.44\,pp above the
student's baseline, the R34 logit distribution may carry little signal beyond
what cross-entropy training already provides, whereas intermediate
representations still differ enough to transfer. The distributional evidence
is consistent with this: the R34 teacher's soft targets are the closest of
all teachers to its student (lowest KL@4 of any Logit-KD cell,
Table~\ref{tab:r34r18}). We continue to flag this as a hypothesis rather
than an established mechanism.

\paragraph{Fidelity Decouples from Accuracy.}
If R34 students absorbed more ``dark knowledge,'' one might expect them to
agree more with their teachers. They do not: top-1 agreement is statistically
flat across all eight KD cells (95.3--95.5\%,
Table~\ref{tab:summary}), despite the accuracy gains concentrating in two of
them. This echoes the central observation of Stanton et
al.~\cite{stanton2021does} that distillation can improve generalization
without producing high-fidelity matching, and it cautions against reading
our capacity result as ``R34 students imitate better'' --- the mechanism, if
real, does not manifest in argmax agreement. A second, unexpected fidelity
pattern does emerge: Feature-KD students end up uniformly closer to their
teacher's output distribution at $T{=}1$ ($D_{KL}$ 0.14--0.19) than Logit-KD
students (0.21--0.28) in every pair --- despite never observing teacher
logits, and despite Logit-KD directly optimizing a (temperature-softened)
KL objective. At the training-relevant temperature $T{=}4$ the comparison is
mixed (Logit-KD is closer on R18 pairs, Feature-KD on R34 pairs). We report
this as an observation; a calibration analysis would be needed to interpret
it.

\paragraph{Effect Sizes and Statistical Discipline.}
Two of eight Welch tests reach $p<0.05$, and none survives Bonferroni
correction across the eight comparisons ($\alpha{=}0.00625$); individual
$p$-values should therefore be read as weak evidence. The case for the
capacity pattern rests instead on its internal consistency: Feature-KD
exceeds Logit-KD in 4/4 pairs, R34 gains exceed R18 gains in every
method-matched contrast, and the two nominally significant cells are exactly
the two the capacity hypothesis predicts. We regard this as suggestive,
directionally consistent evidence --- not proof --- and note that all
observed KD effects are an order of magnitude smaller than architectural
effects (Table~\ref{tab:stem}).

\paragraph{Implementation Correctness, Revisited.}
Version 1 of this study attributed Feature-KD's apparent underperformance to
a gradient-clipping bug. The controlled re-run of Section~\ref{sec:impl}
falsifies that attribution: with projection gradient norms peaking at $0.21$
against a clipping threshold of $1.0$, the bug was inert, and bugged and
corrected variants are statistically identical. The correction that
\emph{did} matter was methodological: replacing test-set selection with
validation-based selection moved every baseline down by roughly $0.2$\,pp
and shrank the largest KD gain from $+0.30$ to $+0.21$\,pp. The lesson we
draw is sharper than v1's: in small-effect-size regimes, finding a bug is
easy, but attributing an effect to it requires a controlled re-run ---
otherwise protocol artifacts masquerade as implementation effects. Results
comparing distillation methods should be interpreted with caution unless
both implementation \emph{and} selection protocol are fully disclosed.

\paragraph{Architecture Dominates KD.}
The CIFAR-specific stem modification is worth $+5.50$ to $+7.15$\,pp
(Table~\ref{tab:stem}) --- more than $25\times$ the largest KD gain observed
in this study. Input-architecture alignment precedes distillation in the
hierarchy of design decisions, and the effect is largest for the smallest
model (R18), the regime where KD is most often deployed.

\section{Limitations}

\paragraph{Dataset Scope.}
All experiments are conducted on CIFAR-10, a 10-class classification benchmark
with $32{\times}32$ images. This choice was deliberate: CIFAR-10 provides a
controlled, low-compute setting that allows us to run systematic ablations (a
full selection grid $\times$ four teacher-student pairs, with five-seed final
runs) without the confounding effects of dataset scale, resolution, and class
imbalance. Isolating architectural correctness and implementation bugs as
sources of variance is substantially easier at this scale. The trade-off is
reduced generalisability: whether the student capacity effect holds on
CIFAR-100, ImageNet, or non-classification tasks remains an open question. We
defer this to future work.

\paragraph{Remaining Confounds and Scope.}
Although the five-seed protocol and a fourth pair strengthen the evidence,
several limitations remain: the study uses a single dataset and a single
Feature-KD variant; student capacity is partially confounded with block
family (BasicBlock vs.\ Bottleneck); only two student sizes are examined; and
the 100-epoch student schedule is shorter than the 240-epoch protocol common
in CIFAR distillation benchmarks, which may affect absolute gains
\cite{beyer2022good}. Results should be interpreted as directional, and
replication on additional pairs and datasets is encouraged.

\section{Conclusion}

We presented a systematic study of knowledge distillation across four
teacher-student capacity pairs on CIFAR-10, comparing Logit-KD and
Feature-KD under a leakage-free protocol: validation-based selection,
five-seed final runs, and fidelity metrics alongside accuracy.

Our central findings are fourfold. First, student capacity, not teacher
scale or teacher-student accuracy gap, is where distillation gains
concentrate: the only significant gains occur for R34 students under
Feature-KD, and doubling teacher parameters at fixed teacher accuracy
(R50 vs.\ R101) leaves the gain unchanged. Given the scope of this study ---
two student sizes, one dataset, effects that do not survive multiple-testing
correction individually --- we treat this as a directionally consistent
finding rather than a definitive claim. Second, a simple Feature-KD baseline
matches or outperforms Logit-KD in all four pairs, and its students land
closer to the teacher's $T{=}1$ output distribution without ever observing
teacher logits. Third, fidelity decouples from accuracy: teacher-student
agreement is flat across all pairs, reinforcing that distillation gains need
not reflect closer imitation \cite{stanton2021does}. Fourth, architectural
correctness dominates: the CIFAR stem modification is worth up to
$+7.15$\,pp, more than $25\times$ any KD effect measured here.

This version also corrects its predecessor in two ways we believe are worth
stating plainly: the gradient-clipping bug blamed in v1 for Feature-KD's
underperformance had, on controlled re-run, no measurable effect; and v1's
larger gains are explained by test-set hyperparameter selection. Small-gain
distillation research is exactly the regime where such protocol artifacts
are largest relative to the effects being measured.

Future work will test whether the student-capacity pattern holds on
CIFAR-100 and ImageNet, across more than two student sizes and disentangled
from block family, and under stronger distillation objectives such as CRD
\cite{tian2020contrastive}, ReviewKD \cite{chen2021distilling}, DKD
\cite{zhao2022decoupled}, and DIST \cite{huang2022stronger}.



\begin{thebibliography}{99}

\bibitem{beyer2022good}
Beyer, L., Zhai, X., Royer, A., Markeeva, L., Anil, R., \& Kolesnikov, A.
(2022).
Knowledge Distillation: A Good Teacher Is Patient and Consistent.
\textit{CVPR}.

\bibitem{chen2021distilling}
Chen, P., Liu, S., Zhao, H., \& Jia, J. (2021).
Distilling Knowledge via Knowledge Review.
\textit{CVPR}.

\bibitem{cho2019efficacy}
Cho, J.~H., \& Hariharan, B. (2019).
On the Efficacy of Knowledge Distillation.
\textit{ICCV}.

\bibitem{he2016deep}
He, K., Zhang, X., Ren, S., \& Sun, J. (2016).
Deep Residual Learning for Image Recognition.
\textit{CVPR}.

\bibitem{hinton2015}
Hinton, G., Vinyals, O., \& Dean, J. (2015).
Distilling the Knowledge in a Neural Network.
\textit{NeurIPS Workshop on Deep Learning}.

\bibitem{huang2022stronger}
Huang, T., You, S., Wang, F., Qian, C., \& Xu, C. (2022).
Knowledge Distillation from a Stronger Teacher.
\textit{NeurIPS}.

\bibitem{mirzadeh2020improved}
Mirzadeh, S.~I., Farajtabar, M., Li, A., Levine, N., Matsukawa, A., \&
Ghasemzadeh, H. (2020).
Improved Knowledge Distillation via Teacher Assistant.
\textit{AAAI}.

\bibitem{romero2015fitnets}
Romero, A., Ballas, N., Kahou, S.~E., Chassang, A., Gatta, C., \& Bengio, Y.
(2015).
FitNets: Hints for Thin Deep Nets.
\textit{ICLR}.

\bibitem{stanton2021does}
Stanton, S., Izmailov, P., Kirichenko, P., Alemi, A.~A., \& Wilson, A.~G.
(2021).
Does Knowledge Distillation Really Work?
\textit{NeurIPS}.

\bibitem{tian2020contrastive}
Tian, Y., Krishnan, D., \& Isola, P. (2020).
Contrastive Representation Distillation.
\textit{ICLR}.

\bibitem{zagoruyko2017paying}
Zagoruyko, S., \& Komodakis, N. (2017).
Paying More Attention to Attention: Improving the Performance of Convolutional
Neural Networks via Attention Transfer.
\textit{ICLR}.

\bibitem{zhao2022decoupled}
Zhao, B., Cui, Q., Song, R., Qiu, Y., \& Liang, J. (2022).
Decoupled Knowledge Distillation.
\textit{CVPR}.

\end{thebibliography}
\end{document}